\documentclass[conference]{article}

\usepackage{url}
\usepackage{graphicx}
\usepackage[cmex10]{amsmath}
\usepackage[noend,ruled,noline,linesnumbered]{algorithm2e}

\usepackage[left = .56in, top=.56in,right=.56in,bottom=0.56in,nohead,nofoot]{geometry}
\usepackage{setspace}

\begin{document}
\title{Stochastic Local Search for Pattern Set Mining}



\author{Muktadir Hossain, Tajkia Tasnim, Swakkhar Shatabda and Dewan M Farid\\
Department of Computer Science and Engineering, United International University\\House \# 80, Road \# 8A, Dhanmondi, Dhaka-1209, Bangladesh\\Email: muktadir00@gmail.com, tajkiamim@gmail.com, swakkhar@cse.uiu.ac.bd, dewanfarid@cse.uiu.ac.bd}

%


\maketitle
\doublespacing

\begin{abstract}
Local search methods can quickly find good quality solutions in cases where systematic search methods might take a large amount of time. Moreover, in the context of pattern set mining, exhaustive search methods are not applicable due to the large search space they have to explore. In this paper, we propose the application of stochastic local search to solve the pattern set mining. Specifically, to the task of concept learning. We applied a number of local search algorithms on a standard benchmark instances for pattern set mining and the results show the potentials for further exploration.
\end{abstract}

\section{Introduction\label{secIntro}}
There have been growing interest in the field of \textit{pattern set mining} in stead of pattern mining in the recent years \cite{bringmann2010mining}. One of the most important task in pattern set mining is to find a particular set of patterns in data that successfully partitions the dataset and discriminates the classes from one another \cite{guns2011declarative}. This is called \textit{concept learning} task in the literature \cite{guns2011declarative}. Such pattern sets are desirable incase of selecting a particular small set of patterns from a large dataset, where traditional pattern mining algorithms fail to produce good results.

In the concept learning task, we are given a set of classes and a set of patterns or features. The task is to select a small set of the features or the patterns so that the classification accuracy is maximized. This is a complex combinatorial optimization task. Most of the algorithms that are applied to solve the problem earlier in the literature are mostly exhaustive or greedy in nature \cite{guns2011declarative}. Declarative frameworks like constraint programming methods  \cite{guns2011declarative,guns2011itemset} have gained some significant success. However, with increasing problem size, these methods struggle to produce good quality solutions within a short period of time. On the other hand, local search methods in general can quickly find good quality solutions and have been very effective to find satisfactory results for many combinatorial optimization problems.

In this paper, we propose to apply a large variety to local search algorithms to solve the concept learning task in the context of pattern set mining. The set of algorithms that we use includes random walk, random valid walk, hill climbing, hill climbing with restart and genetic algorithm. We performed our experiments to solve standard benchmark datasets extensively used by the researchers in the literature. The key contributions in the paper are as follows:
\begin{itemize}
\item Demonstrate the overall strength of stochastic local search methods solving the pattern set mining task.
\item Perform a comparative analysis of various local search algorithm and analysis of their relative strength compared with each other.
\end{itemize}

The rest of the paper is organized as follows: Section~\ref{secPreli} describes the concept learning task in the context of pattern sets mining; Section~\ref{secRel} reviews the related work; Section~\ref{secMeth} describes the algorithms used; Section~\ref{secEx} discusses and analyzes the experimental results; and finally, Section~\ref{secCon} presents our conclusions and outlines our future work.
\section{Preliminaries\label{secPreli}}
In this section, we briefly describe the problem model that we use. We adopt the itemset mining \cite{guns2011itemset} setting used earlier by Guns et al. in \cite{guns2011declarative}. In the concept learning task we are given a database of transactions, $\mathcal{D}$. Which is a binary matrix, meaning all items are either 0 or 1. The columns of this binary matric corresponds to a set of patterns or items, $\mathcal{I}$ and the rows corresponds to actual data items or transactions. The set of transactions is denoted by $\mathcal{T}$. For example consider the dataset given in Table~\ref{table1}. Here, we have four items in the itemset, $\mathcal{I}=\{A,B,C,D\}$ and five transactions in the transaction set, $\mathcal{T}=\{t_1,t_2,t_3,t_4,t_5\}$.
\begin{table}
\caption{A small example dataset containing four items and five transactions.\label{table1}}
\begin{center}
\begin{tabular}{clccccc}
\hline
Transaction&&&&&\\
Id&ItemSet&A&B&C&D&Class\\
\hline
$t_1$&\{A,B,D\}&1&1&0&1&+\\
$t_2$&\{B,C\}&0&1&1&0&+\\
$t_3$&\{A,D\}&1&0&0&1&+\\
$t_4$&\{A,C,D\}&1&0&1&1&-\\
$t_5$&\{B,C,D\}&0&1&1&1&-\\
\hline
\end{tabular}

\end{center}
\end{table} 
The itemsets or pattern sets and the transaction sets are generally represented by binary vectors. The \textit{coverage} $\varphi_{\mathcal{D}}(I)$ of an itemset $I$ consists of all transactions in which the itemset occurs:
$$\varphi_{\mathcal{D}}(I)=\{t \in \mathcal{T}|\forall i \in I: \mathcal{D}_{ti}=1\}$$

For example, given an itemset, $I={A,D}$, it is represented as $\langle 1,0,0,1 \rangle$ and the the coverage is $\varphi{\mathcal{D}}(I)=\{t_1,t_3,t_4\}$ which is represented by $\langle 1,0,1,1,0 \rangle$. Support of the itemset is $Support_{\mathcal{D}}(I)= 3$. Where, Support of an itemset is the size of its coverage set, $Support_{\mathcal{D}}(I)=|\varphi{\mathcal{D}}(I)|$.

Any item set $I$ is \textit{closed} over $T$, if  no other supersets of this itemset covers the same transaction set $T$ as $I$. Formally, closedness $\psi(I,T)$ is defined as the following constraint:
$$\psi(I,\mathcal{T})=1 \Leftrightarrow (\forall i\in \mathcal{I}:I_i = 1 \Leftrightarrow \underset{t\in \mathcal{T}}{\bigwedge}(\mathcal{D}_{ti}=1 \lor T_t=0)) $$

For example, the itemset $I_1={A,D}$ is closed and the itemset, $I={A,C}$ is not closed. Each of the transactions in the dataset are associated with two distinct classes: positive transactions $\mathcal{T^+}$ and negative transactions $\mathcal{T^-}$. Now, we can define the \textit{accuracy} of an itemset or patternset which is defined by its coverage set $T$ as following:
$${accuracy}(T)= \underset{t\in \mathcal{T^+}}{\sum}{T_t}-\underset{t\in \mathcal{T^-}}{\sum}{T_t}$$

In pattern set mining, we are interested to find $k-$pattern sets \cite{guns2013k}. A $k-$pattern set $\Pi$ is a set of $k$ tuples, each of type $\langle I^p,T^p \rangle$. The pattern set is formally defined as following:
$$\Pi=\{\pi_1,\cdots,\pi_k\}, \text{ where, } \forall p =1,\cdots,k: \pi_p=\langle I^p,T^p \rangle$$ 

The concept learning task is to find a set of itemsets or patterns that together cover as many positive transactions as possible,
while covering only few negative transactions. This can be formalized as the following optimization problem:

 \[
\begin{array}{l}
\underset{\Pi,T}{\textsf{ maximize }} { accuracy}(T)\\
\text{ where, }\\
\forall t \in \mathcal{T}: T_t=(\vee_{p=1\cdots k}T^p_t) \\
\forall \pi \in \Pi: \varphi_{\mathcal{D}}(I^p)=T^p \\

\forall \pi \in \Pi: \psi(I,\mathcal{T}^+)=1\\
\end{array}
\]

In this paper, we attempt to solve this problem using different types of stochastic local search algorithms: random walk, random valid walk, hill climbing, hill climbing with restart and genetic algorithm.

\section{Related Works\label{secRel}}
 The frequent itemset mining task was first proposed in \cite{agrawal1993mining} followed by a large number of algorithms and their application is various fields. In the the pattern mining problem \cite{hand2002pattern}, we are interested to find patterns in the dataset that are correlated \cite{morishita2000transversing}, discriminative \cite{cheng2007discriminative}, contrast \cite{novak2009supervised}, diverse \cite{ruckert2007optimizing} etc. A large variety of algorithms has been proposed to solve the pattern set mining in the last decade \cite{bringmann2010mining}.  

Recent advancements in the field shows the use of declarative programming frameworks i.e. constraint programming \cite{guns2011itemset}. State-of-the-art constraint solvers i.e. COMET \cite{hentenryck2009constraint} are used to solve the pattern set mining and related tasks \cite{guns2011declarative}. It opens a huge research area by combining data mining or machine learning tasks with constraint programming \cite{de2012declarative,guns2013miningzinc}. However, most of these methods use systematic search methods to search and shrink the solution space. Most of the systematic search algorithms are exhaustive in nature and require huge amount of time and resources to solve the problem efficiently. Where as greedy methods does not guarantee optimality. The necessity of efficient global optimization algorithms and effective search heuristics for such problems is higher than ever. 

In \cite{ruckert2007optimizing}, stochastic search algorithms are applied to solve a related problem to find diverse pattern sets. However, Guns et al. \cite{guns2011declarative} applied large neighborhood search, using constraint programming platform to solve both of the concept learning task and the diverse pattern set mining problem.
 
\section{Our Approach\label{secMeth}}
In this section, we describe the algorithms that we have implemented to solve the concept learning task. The basic framework contains two important functions that calculates the coverage and checks the closedness or maximal property of the itemsets or patternsets.

\begin{algorithm}
\DontPrintSemicolon
\caption{getCoverage(ItemSet $I$, TransactionSet $\mathcal{T}$)\label{algoGC}}
$coverage=[0,\cdots, 0]$\;
\For{each transaction t $\in$ $\mathcal{T}$}{
\If {all items in $I$ appears in $t$}
{
	$coverage[t]=1$
}
}
\Return $coverage$

\end{algorithm}

The \textit{getCoverage} algorithm is given in Algorithm~\ref{algoGC}. It simply searches for occurrences of items in an itemset in all transactions in the dataset and returns them all. The next important function is the \textit{isClosed} function that verifies the maximal property of an itemset. The algorithm is pictured in Algorithm~\ref{algoIC}.

\begin{algorithm}
\DontPrintSemicolon
\caption{isClosed(ItemSet $I$,Database $\mathcal{D}$)\label{algoIC}}
$s_0 = Support_{\mathcal{D}}(I)$\;

\For {each item $i \in \mathcal{D}$ }{
\If{$i \notin I$}
{
	$I_n=I\cup\{i\}$\;
	$s_n=Support_{\mathcal{D}}(I_n)$\;
	\If{$s_n \ge s_0$}
	{
		\Return false\;
	}
}
}
\Return true\;
\end{algorithm}

The calculation of accuracy is done by calculating the number of positive transactions, covered by any of the coverage sets of the itemsets in the pattern set and subtracting the number of negative transactions covered by any of the coverage sets of the itemsets, in the pattern set. The algorithm for calculating accuracy in Algorithm~\ref{algoCA}.

\begin{algorithm}
\DontPrintSemicolon
\caption{getAccuracy(PatternSet $\Pi$, Database $\mathcal{D}$)\label{algoCA}}
$\mathcal{C}^+$: set of positive coverages \; 
$\mathcal{C}^-$: set of negative coverages \;
\For{each itemset $I \in \Pi$}{
$\mathsf{\mathcal{C}^+[i]=\mathsf{getCoverage}(I,\mathcal{T}^+)}$ \;
$\mathsf{\mathcal{C}^-[i]=\mathsf{getCoverage}(I,\mathcal{T}^-)}$ \;}

$posN=0$\;
\For{each transaction $t \in \mathcal{T}^+$}
{
	\For{each coverage $c \in \mathcal{C}^+$ }
	{
		\If{$c[t]==1$}
		{
			$posN++$\;
			break\;
		} 
	}
}
$negN=0$\;
\For{each transaction $t \in \mathcal{T}^-$}
{
	\For{each coverage $c \in \mathcal{C}^-$ }
	{
		\If{$c[t]==1$}
		{
			$negN++$\;
			break\;
		} 
	}
}
\Return $posN-negN$\;
\end{algorithm}

Rest of the section describes the algorithms that we have implemented on this setup using these functions.

\subsection{Random Walk}
The random walk algorithm is similar to Monte Carlo simulations in nature. The pseudo-code for the algorithm is given in Algorithm~\ref{algoRW}.
\begin{algorithm}
\DontPrintSemicolon
\caption{RandomWalk()\label{algoRW}}
$maxAccuracy = -\infty$ \;
$\Pi^*=\phi$\;
\While{timeout}{
$\Pi=$ randomly create a k-itemset\;
$isMaximal=true$\;
\For {each itemset $I \in \Pi$}{
\If{$\mathsf{isClosed}{(I,\mathcal{T^+})}==false$}
{
$isMaximal=false$\;
break\;
}
}
\If{$isMaximal==true$}{

$accuracy = \mathsf{getAccuracy}(\Pi,\mathcal{D})$\;
\If{$accuracy > maxAccuracy$}
{
$maxAccuracy=accuracy$\;
$\Pi^*=\Pi$\;

}
}
}
\Return $\Pi^*$\;
\end{algorithm}
This algorithm continuously updates the maximum accuracy and the best solution $\Pi^*$ and then returns the best solution found before the algorithm times out.

\subsection{Random Valid}
The next algorithm we call random valid. It starts by initially generating a valid patterns set. A pattern set is valid if the itemsets are all maximal. The termination criteria is similar to the previous algorithm random walk. In this version, the algorithm make some random changes in the current solution and accepts the new solution only if the resulting pattern set is valid in each iteration. Our random valid algorithm is given in Algorithm~\ref{algoRV}.

\begin{algorithm}
\DontPrintSemicolon
\caption{RandomValid()\label{algoRV}}

$\Pi =$ generate a valid pattern set with $k$ items\;
$\Pi^*=\Pi$\;
$maxAccuracy = \mathsf{getAccuracy}(\Pi,\mathcal{D})$\;

\While{timeout}{
$\Pi_t$= make random change in $\Pi$\;
\If{$\textsf{isValid}(\Pi_t,\mathcal{D})$}{
$accuracy = \mathsf{getAccuracy}(\Pi,\mathcal{D})$\;
\If{$accuracy > maxAccuracy$}
{
$maxAccuracy=accuracy$\;
$\Pi^*=\Pi_t$\;

}
$\Pi=\Pi_t$\;
}
}
\Return $\Pi^*$
\end{algorithm}

\subsection{Hill Climbing}
The hill climbing algorithm is similar to the random valid algorithm. But it differs in how the next candidate solution is accepted at Lines 10-11. The new candidate solution is accepted only if it does not decrease the accuracy of the current candidate pattern set. The algorithm is given in Algorithm~\ref{algoHC}.
\begin{algorithm}
\DontPrintSemicolon
\caption{HillClimbing()\label{algoHC}}
$\Pi =$ generate a valid pattern set with $k$ items\;
$maxAccuracy = \mathsf{getAccuracy}(\Pi,\mathcal{D})$\;

\While{timeout}{
$\Pi_t$= make random change in $\Pi$\;
\If{$\textsf{isValid}(\Pi_t,\mathcal{D})$}{
$accuracy = \mathsf{getAccuracy}(\Pi,\mathcal{D})$\;
\If{$accuracy \ge maxAccuracy$}
{
$maxAccuracy=accuracy$\;
$\Pi=\Pi_t$\;
}
}
}
\Return $\Pi^*$
\end{algorithm}

\subsection{Hill Climbing with Restart}
The hill climbing algorithm pictured in Algorithm~\ref{algoHC} is greedy in nature as it only accepts non-decreasing steps, in terms of accuracy. However, such strategy quickly leads the algorithm into local minima and cannot improve further. To tackle this situation, we propose the next algorithm which we call hill climbing with restart. It denoted by $HC+Restart$ in the rest of the paper. The pseudo-code for $HC+Restart$ is given in Algorithm~\ref{algoHCWR}. This algorithm is similar to that of hill climbing but it keeps track of the non improving steps and then it restarts the search after the number of non improving steps, crosses the limit of a parameter $threshold$. The value of $threshold$ was kept 100 for these experiments. 

\begin{algorithm}
\DontPrintSemicolon
\caption{HillClimbingWithRestart()\label{algoHCWR}}
$\Pi =$ generate a valid pattern set with $k$ items\;
$\Pi^*=\Pi$\;
$maxAccuracy = \mathsf{getAccuracy}(\Pi,\mathcal{D})$\;
$nonImprovingSteps=0$\;
\While{timeout}{
$\Pi_t$= make random change in $\Pi$\;
\If{$\textsf{isValid}(\Pi_t,\mathcal{D})$}{
$accuracy = \mathsf{getAccuracy}(\Pi,\mathcal{D})$\;
\If{$accuracy > maxAccuracy$}
{
$maxAccuracy=accuracy$\;
$\Pi^*=\Pi_t$\;
$nonImprovingSteps=0$\;
}
\Else
{
	$nonImprovingSteps++$\;
}
\If{$nonImprovingSteps \ge threshold$}
{
	$\Pi =$ generate a valid pattern set with $k$ items\;
	$nonImprovingSteps=0$\;
}
\Else
{
	$\Pi=\Pi_t$\;
}
}
}
\Return $\Pi^*$
\end{algorithm}

\subsection{Genetic Algorithm}
The single point search methods often starts from a solution in the search space and do not have much chance of divergence. Which is needed to cover the huge search space for global optimization problem. For this reason, multi-point search algorithms are preferred. They keep a number of solutions in the population and runs single point search in parallel. Moreover, combinations of the individuals in the population helps the search to create high quality solutions that resembles the  natural process of evolution of species. One such algorithm is genetic algorithms. The genetic algorithm that we propose here keeps a population of pattern sets and then iteratively recombines and mutates the individuals to create new solutions. The pseudo-code for our genetic algorithm is given in Algorithm~\ref{algoG}.

\begin{algorithm}
\DontPrintSemicolon
\caption{Genetic Algorithm()\label{algoG}}
$p:$ population size\;
$\mathcal{P}=$ generate $p$ valid pattern sets\;
\While{$timeout$}
{
	$childCount=0$\;
	$\mathcal{P}_c=\{\}$\;
	\While{$childCount < p$}
	{
		$\Pi_1=\mathsf{selectParentAtRandom}(\mathcal{P})$\;
		$\Pi_2=\mathsf{selectParentAtRandom}(\mathcal{P})$\;
		$\Pi_c=\mathsf{crossover}(\Pi_1,\Pi_2)$\;
		\If{$\mathsf{isValid}(\Pi_c)$}
		{
			$\mathcal{P}_c=\mathcal{P}_c \cup \{\Pi_c\}$\;
			$childCount++$\;
		}
	}
	$\mathcal{P}_m=\{\}$\;
	\For{each $\Pi \in \mathcal{P}_c$ }
	{
		\While{true}
		{
			$\Pi_m=$ make random change in $\Pi$\;
			\If{$\mathsf{isValid}(\Pi_m)$}
				{
					$\mathcal{P}_m=\mathcal{P}_m \cup \{\Pi_m\}$\;
						break\;
				}
		}
	}
	$\mathcal{P}=\mathsf{selectBest}(\mathcal{P}\cup \mathcal{P}_c\cup \mathcal{P}_m)$
}
\Return $globalBest$\;
\end{algorithm}

In our genetic algorithm, the population is initialized by a number of valid pattern sets. In each generation of the genetic algorithm, the algorithm goes through two phases: crossover and mutation. In the crossover phase, two parent individuals are choosen randomly from the population and a child individual in created. This new individual is created using the one point crossover operation. After the validity of the individual checking, only a valid solution is added to the crossover population. This process iterates until the crossover population size is equal to the population size. Then the mutation phase begins. From each of the valid individuals in the crossover population a valid individual is created and added to the mutation population list. At the end of the mutation phase best solutions from these three population lists are selected for the next generation. 

\section{Experimental Results\label{secEx}}
We have implemented the algorithms in Ruby language and have run our experiments on an Intel core $i3$ 2.40 GHz machine with 6 GB ram running 32bit Ubuntu 14.04 operating system.
\subsection{Dataset}
The benchmark datasets that we use in this paper are taken from UCI Machine Learning repository~\cite{frank2010uci} and originally used in \cite{guns2011declarative}. The datasets are available to download freely from the website: \url{https://dtai.cs.kuleuven.be/CP4IM/datasets/}. The benchmarks are given in Table~\ref{tableB} with their properties.

\begin{table}[h]
\begin{center}
\caption{Description of benchmark datasets.\label{tableB}}
\begin{tabular}{l|cccc}
\hline
Data set & Items & Transactions & Density & Class Distribution\\
\hline
zoo-1&36&101&44\%&41\%\\
hepatitis&44&137&50\%&81\%\\
lymph&60&148&38\%&55\%\\
heart-cleaveland&95&296&47\%&54\%\\
vote&48&435&33\%&61\%\\
primary-tumor&31&336&48\%&24\%\\

\hline
\end{tabular}
\end{center}
\end{table}
\subsection{Results}
We performed experiments by running different algorithms on each of the benchmark dataset and reported accuracy in Table~\ref{tableM}. Due to the unavailability of COMET, we were unable to compare the search performance with the large neighborhood search used in \cite{guns2011declarative}. Each algorithm were given 10 minutes to finish and the best accuracy found during the runtime is reported in the table. The pattern size setting was varied to see the effect of the size parameter on the performance of the algorithms. The value of the parameter $k$ was varied for the values $\{2,3,4\}$. The best values in each row are shown in bold faced fonts.

\begin{table*}[t]
\caption{Accuracy reported by different algorithms for various datasets with differing sizes of pattern sets $k$.\label{tableM}}
\begin{center}
\renewcommand*{\arraystretch}{1.4}
\begin{tabular}{l|c|cc|cc|cc|cc|cc}
\hline
&Pattern &\multicolumn{5}{c}{Search Algorithm}\\
\cline{3-12}
 &Set Size& \multicolumn{2}{c}{Random Walk} &  \multicolumn{2}{c}{Random Valid} &  \multicolumn{2}{c}{Hill Climbing} &  \multicolumn{2}{c}{HC + Restart} &  \multicolumn{2}{c}{Genetic Algorithm}\\
\cline{3-12}
{Data set}&{k}&{best}&{avg}&{best}&{avg}&{best}&{avg}&{best}&{avg}&{best}&{avg}\\
\hline
zoo-1&2&{100}&{66.87}&{82.13}&{73.93}&{94.15}&{82.74}&{100}&{92.07}&{100}&{98.06}\\
&3&{93.7}&{73.73}&{98.09}&{77.07}&{98.18}&{85.25}&{100}&{94.71}&{100}&{97}\\
&4&{96.92}&{76.97}&{92.17}&{78.19}&{91.97}&{84.4}&{94.76}&{89.71}&{100}&{96.71}\\
\hline
hepatitis&2&{89.17}&{71.7}&{86.06}&{76.43}&{87.19}&{78.18}&{100}&{92.13}&{100}&{97.49}\\
&3&{82.07}&{78.91}&{83.04}&{77.52}&{95.76}&{82.96}&{100}&{91.93}&{100}&{94.7}\\
&4&{89.47}&{75.24}&{87.48}&{77.57}&{100}&{86.52}&{100}&{93.13}&{100}&{96.24}\\
\hline
lymph&2&{81.4}&{76.51}&{100}&{78.35}&{97.4}&{81.94}&{100}&{92.97}&{100}&{96.72}\\
&3&{83.41}&{71.17}&{98.47}&{75.64}&{91.67}&{82.07}&{100}&{90.85}&{100}&{95.04}\\
&4&{85.67}&{72.62}&{86.89}&{82.29}&{89.54}&{80.56}&{92.38}&{90.42}&{100}&{95.7}\\
\hline
heart-cleaveland&2&{96.69}&{77.35}&{84.27}&{75.71}&{100}&{86.3}&{100}&{89.34}&{100}&{96.89}\\
&3&{100}&{82.75}&{90.78}&{70.31}&{97.51}&{85.28}&{97.87}&{85.8}&{100}&{95.21}\\
&4&{98.78}&{76.71}&{96.4}&{78.1}&{97.56}&{83.41}&{100}&{91.38}&{100}&{93.29}\\
\hline
vote&2&{83.41}&{71.41}&{85.21}&{71.44}&{98.42}&{81.78}&{100}&{89.76}&{100}&{94.84}\\
&3&{85.54}&{72.3}&{87.4}&{73.79}&{87.41}&{80.34}&{96.12}&{89.6}&{100}&{93.84}\\
&4&{78.05}&{68.82}&{87.08}&{77.97}&{84.91}&{80.48}&{98.84}&{89.84}&{98.46}&{94.48}\\
\hline
primary-tumor&2&{100}&{80.75}&{88.69}&{72.61}&{85.54}&{78.37}&{96.47}&{87.52}&{100}&{93.25}\\
&3&{88.69}&{72.11}&{78.46}&{72.65}&{87.59}&{79.69}&{97.87}&{86.81}&{100}&{91.41}\\
&4&{82.26}&{66.64}&{84.65}&{70.91}&{89.54}&{82.11}&{98.57}&{87.48}&{100}&{93.14}\\
\hline
\end{tabular}
\end{center}\end{table*}
\subsection{Analysis}

From the results reported in Table~\ref{tableM}, it is evident that genetic algorithms work better than other algorithms for most of the datasets for different pattern sizes. However, in a few cases hill climbing with restart (denoted by HC+Restart in the table) works better. However, the margin is not much higher. The random walk Monte Carlo simulation performs poorly since there not much of intelligent decisions taken during the search. The performance of the random valid search improves since random walk wastes a good amount of time finding valid pattern sets that are closed over the positive transaction set. The hill climbing search quickly gets stuck into the local maximum and once stagnates, it can not improve further due to lack of escaping mechanism. The situation improves when we add random restart strategy along with the hill climbing search. The performance of the algorithms somehow varies with the size of the pattern sets. The situation is pictured in \figurename~\ref{figBar}. In this figure, accuracy of the search algorithms are shown as vertical bars for all the datasets for different pattern set sizes. It is evident from the bar diagrams that the increase in the pattern set size favors the genetic algorithm.

\begin{figure}
\begin{center}
\includegraphics[width=0.4\textwidth]{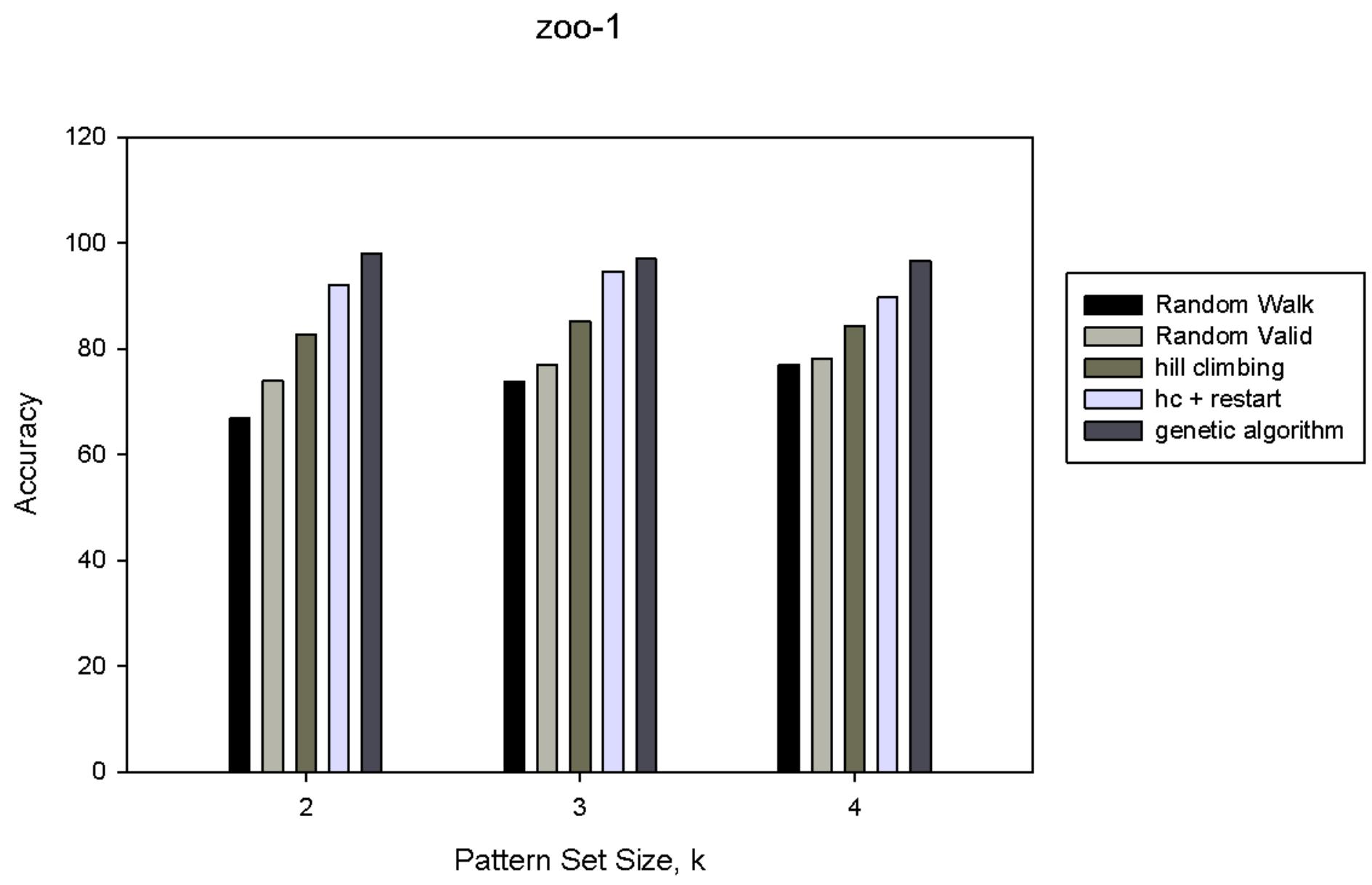}
\caption{Bar diagram showing comparison of accuracy achieved by different algorithms for various sizes of pattern sets, $k=2,3,4$. \label{key1}}
\end{center}
\end{figure}

\begin{figure}
\begin{center}
\includegraphics[width=0.4\textwidth]{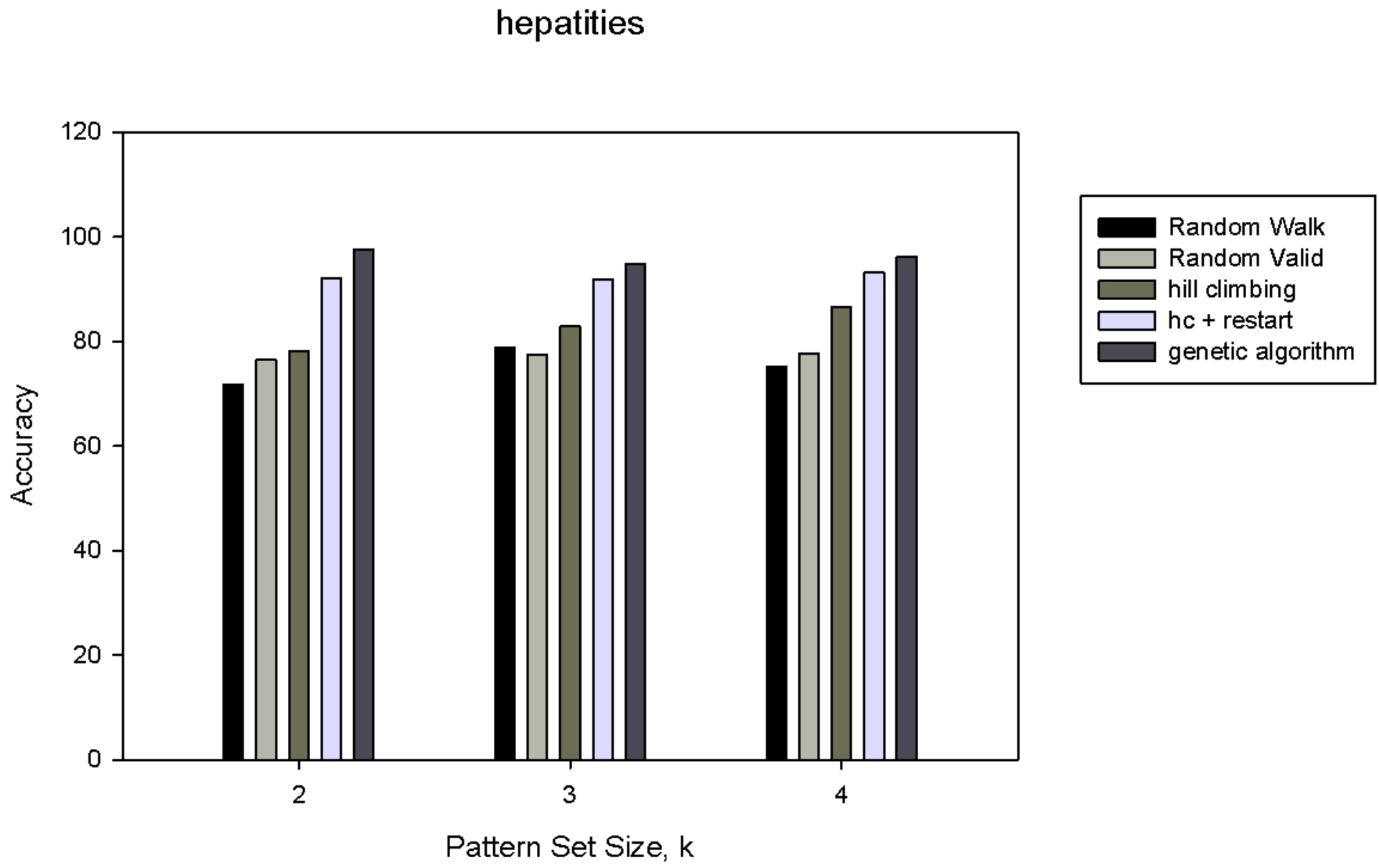}
\caption{Put a title here \label{key2}}
\end{center}
\end{figure}

\begin{figure}
\begin{center}
\includegraphics[width=0.4\textwidth]{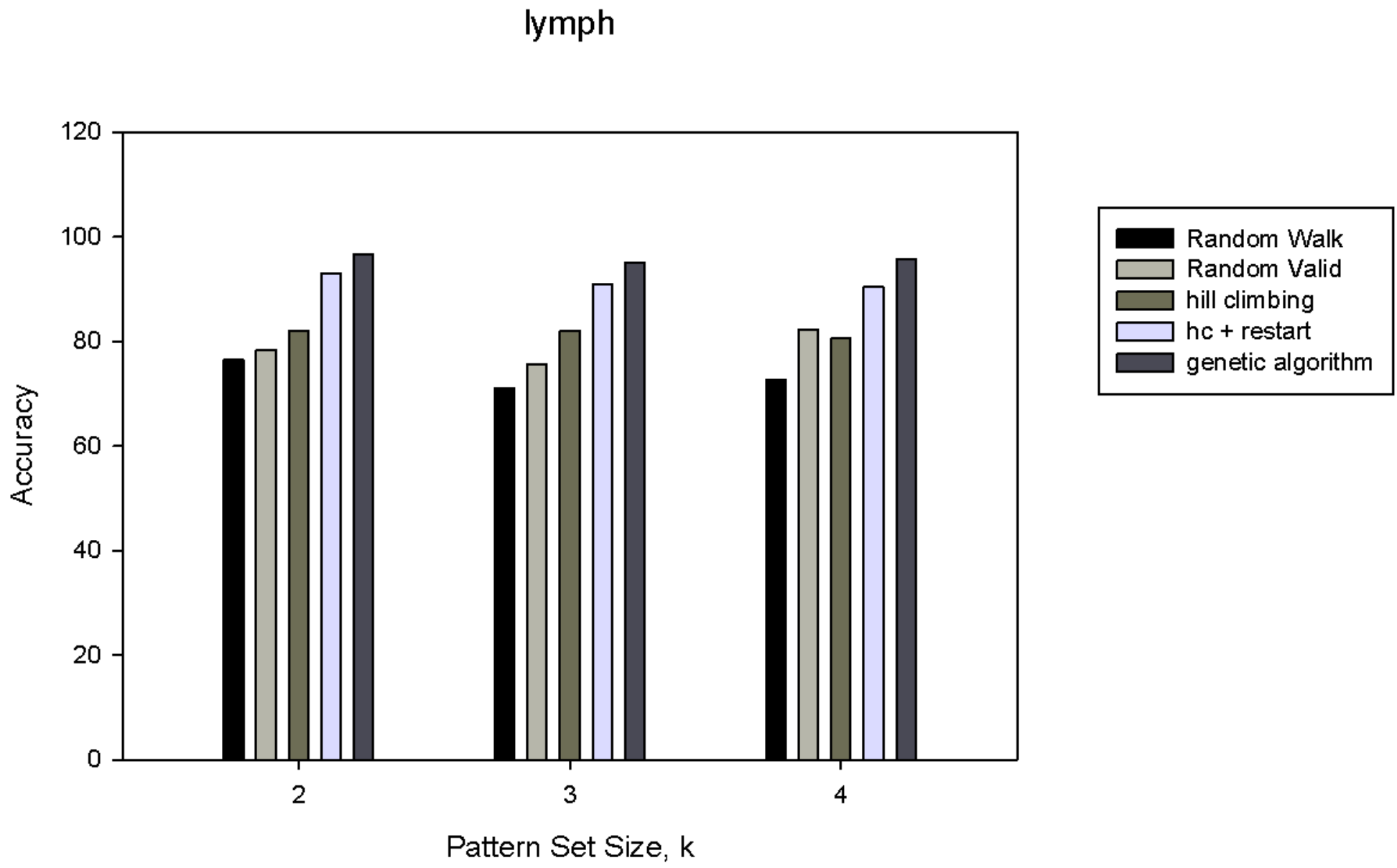}
\caption{Put a title here \label{key3}}
\end{center}
\end{figure}

\begin{figure}
\begin{center}
\includegraphics[width=0.4\textwidth]{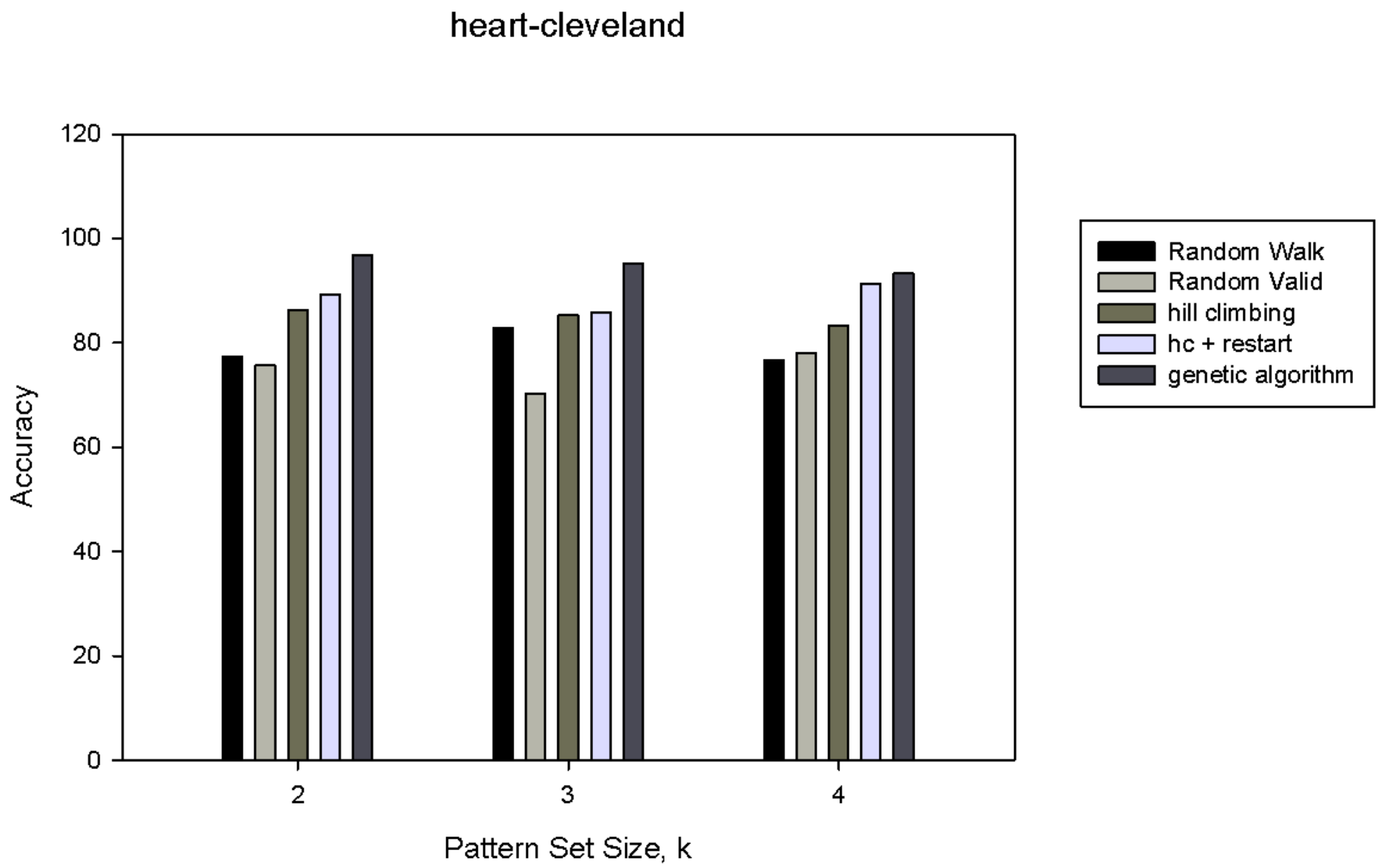}
\caption{Put a title here \label{key4}}
\end{center}
\end{figure}

\begin{figure}
\begin{center}
\includegraphics[width=0.4\textwidth]{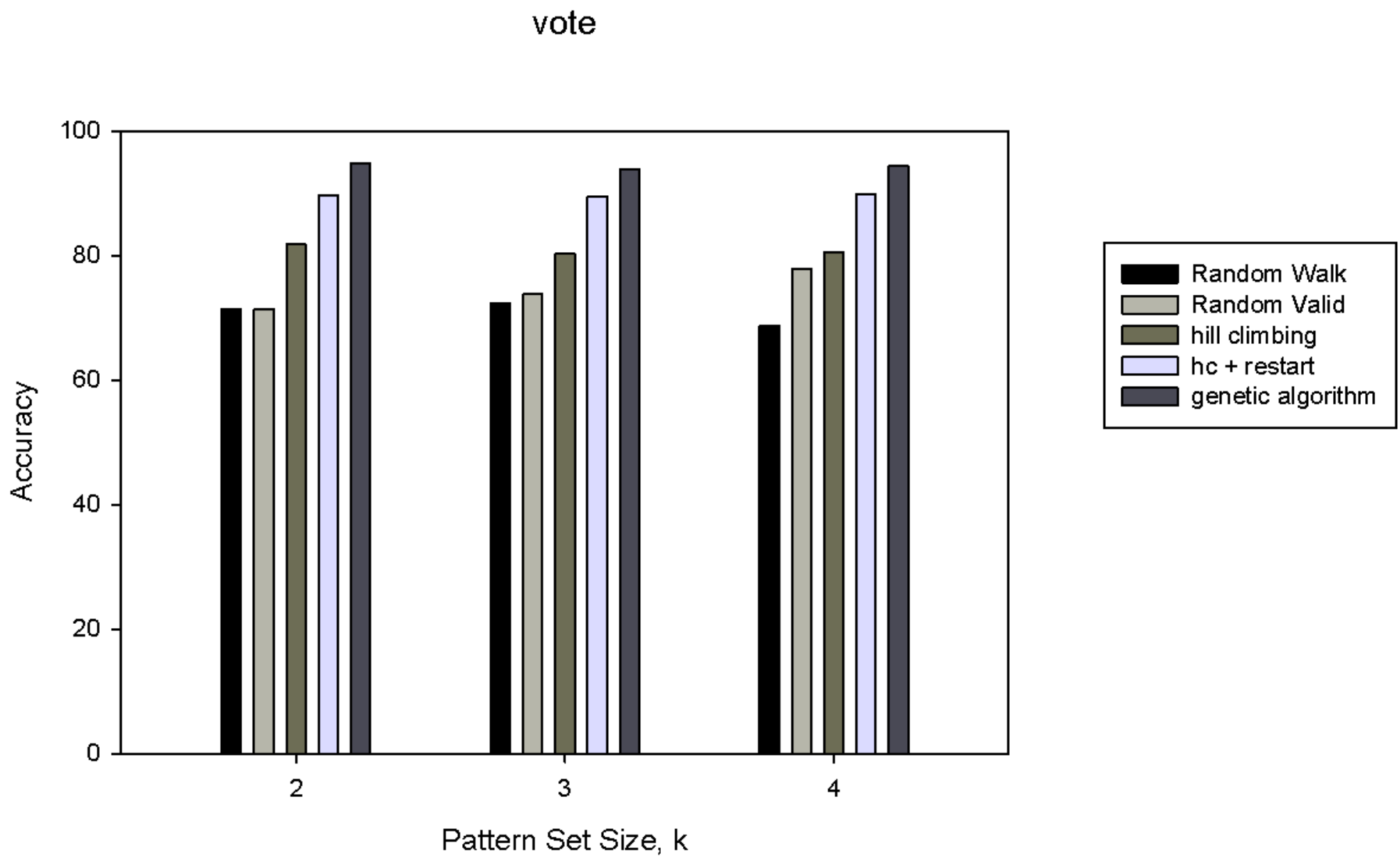}
\caption{Put a title here \label{key5}}
\end{center}
\end{figure}

\begin{figure}
\begin{center}
\includegraphics[width=0.4\textwidth]{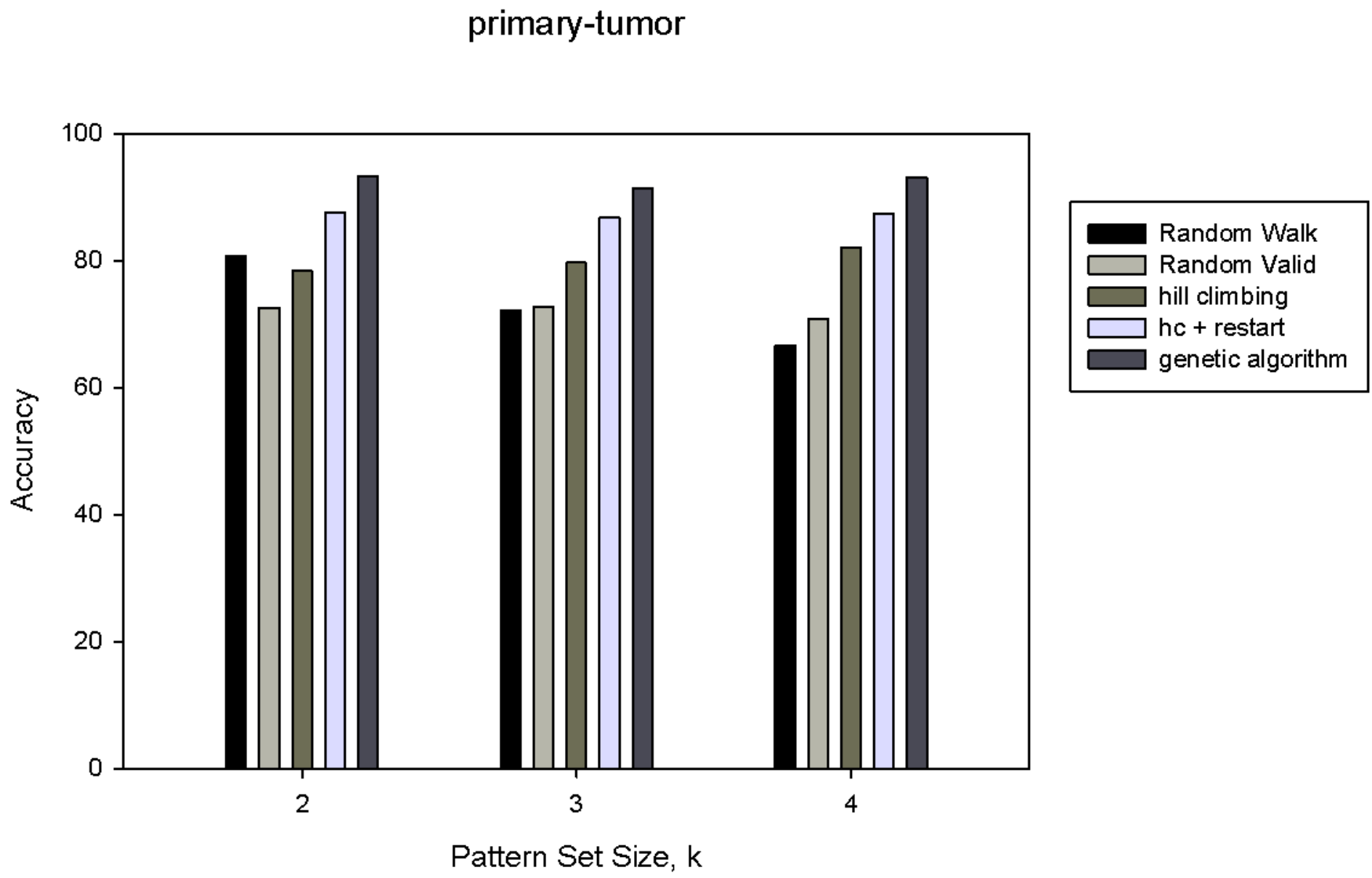}
\caption{Put a title here \label{key6}}
\end{center}
\end{figure}

\begin{figure}
\begin{center}
\includegraphics[width=.4\textwidth]{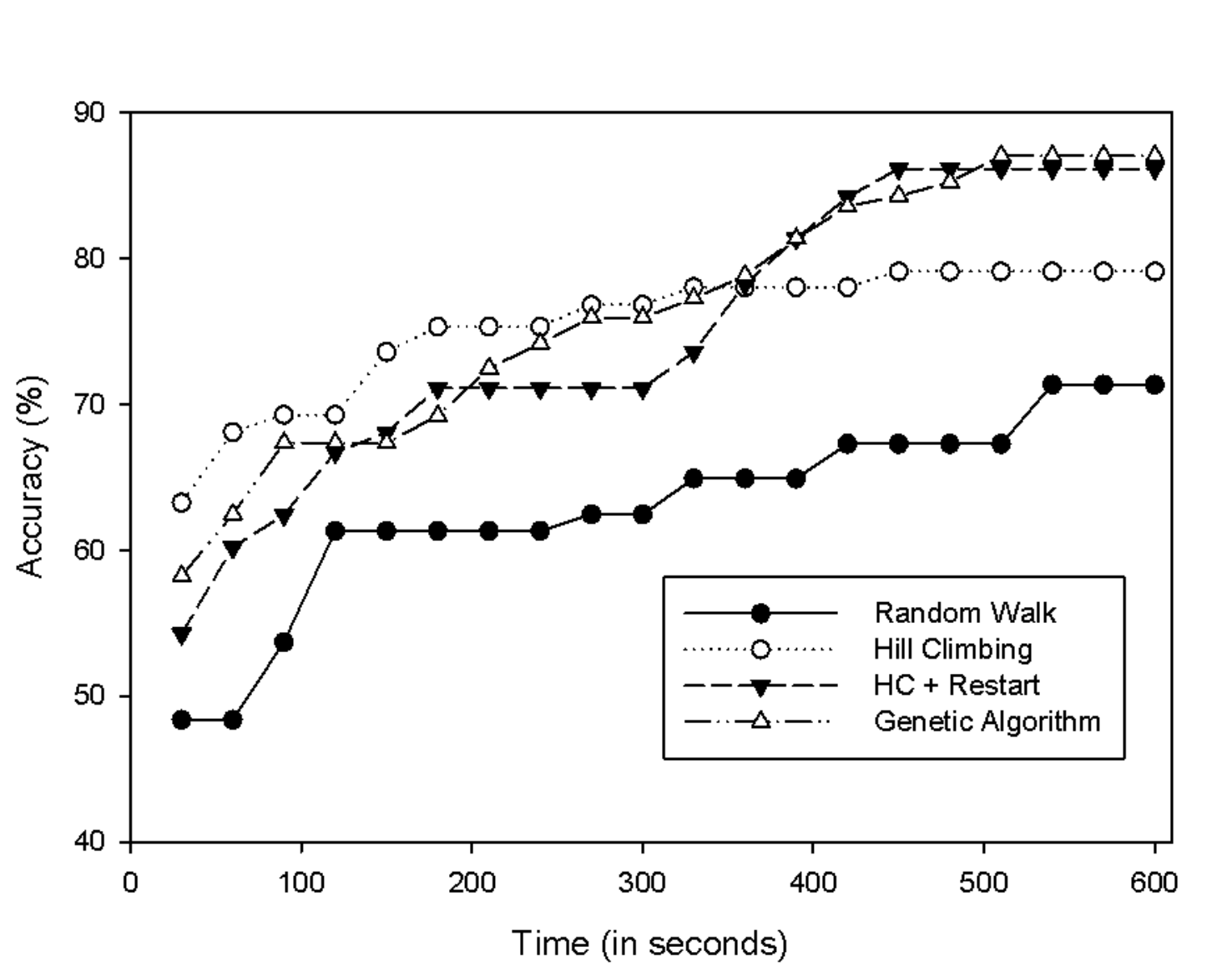}
\end{center}
\caption{Search progress for different algorithm for the hepatitis dataset with size of the pattern size $k=3$.\label{figProg}}
\end{figure}

We also depict the performance of the search algorithms in \figurename~\ref{figProg}. It shows the progress of different search algorithms in 10 minutes for a single run of the \textit{hepatitis} dataset. The random walk algorithm lacking intelligent decision through the search can not improve much. Where as, hill climbing improves very quickly by taking greedy best choice moves. However, if it gets stuck and the improvement is possible with random restarts. Genetic algorithm shows continuous improvements of best accuracy achieved.
\section{Conclusion\label{secCon}}
In this paper, we showed the effectiveness of various stochastic local search algorithms to solve the concept learning task in pattern set mining. Population based algorithms like genetic algorithm shows promising results while local minima escaping strategies like random restart increases the effectiveness of greedy hill climbing search to a great extent. In future, we would like to improve the performance of the search techniques by incorporating further effective strategies within the framework of stochastic local search and solve pattern set mining related problems with realistic datasets.

\bibliographystyle{IEEEtran}
\bibliography{muktadir}

\end{document}